\documentclass[letterpaper, 10 pt, conference]{ieeeconf}

\usepackage{cite}
\usepackage{graphicx}
\usepackage{tcolorbox}
\usepackage{xcolor}
\usepackage[hidelinks]{hyperref}
\usepackage[]{footmisc}
\usepackage{adjustbox}
\usepackage{booktabs,graphicx,makecell,siunitx,eqparbox}
\usepackage{tabularx}
\usepackage{footnote}
\usepackage{balance}
\usepackage{caption}
\makesavenoteenv{tabular}
\usepackage{amssymb}
\usepackage{subfigure}
\usepackage{algorithm}
\usepackage{algorithmic}

\IEEEoverridecommandlockouts  
\usepackage[nopostdot,nogroupskip,nonumberlist]{glossaries}
\makeglossaries

\usepackage{array}
\newcolumntype{L}[1]{>{\raggedright\let\newline\\\arraybackslash\hspace{0pt}}m{#1}}
\newcolumntype{C}[1]{>{\centering\let\newline\\\arraybackslash\hspace{0pt}}m{#1}}
\newcolumntype{R}[1]{>{\raggedleft\let\newline\\\arraybackslash\hspace{0pt}}m{#1}}

\IEEEoverridecommandlockouts                              % For Thanks

\title{\LARGE \bf
LensDFF: Language-enhanced Sparse Feature Distillation for Efficient Few-Shot Dexterous Manipulation
}

\author{Qian Feng$^{*,1,2}$, David S. Martinez Lema$^{*,1,2}$, Jianxiang Feng$^{1,2}$, Zhaopeng Chen$^{1}$, Alois Knoll$^{2}$
\thanks{*: Equal Contributions, \{qian.feng,  david.martinez\}@tum.de.}
\thanks{$^{1}$Agile Robots SE}
\thanks{$^{2}$ TUM School of Information Computation and Technology, Technical University of Munich}
}

\usepackage{booktabs}
\newcommand{\ra}[1]{\renewcommand{\arraystretch}{#1}}

\begin{document}

\maketitle
% \thispagestyle{empty}
% \pagestyle{empty}

%b
% \newglossaryentry{bnn}{name=BNNs, description={Bayesian Neural Networks},first={Bayesian Neural Networks (BNNs)}}
\newglossaryentry{bps}{name=BPS, description={Basis Point Set},first={Basis Point Set (BPS)}}
%c
\newglossaryentry{cgm}{name=CGMs, description={Conditional Generative Models},first={Conditional Generative Models (CGMs)}}
\newglossaryentry{cvae}{name=cVAE, description={Conditional Variational Autoencoder},first={Conditional Variational Autoencoder (cVAE)}}
\newglossaryentry{ci}{name=CI, description={Mutual Cross-Information},first={Mutual Cross-Information (CI)}}
\newglossaryentry{cov}{name=Cov, description={Coverage},first={Coverage (Cov)}}
%d
\newglossaryentry{dof}{name=DoF, description={Degrees of Freedom},first={Degrees of Freedom (DoF)}}
% \newglossaryentry{dlvm}{name=DLVMs, description={Deep Latent Variable Models},first={Deep Latent Variable Models (DLVMs)}}
% \newglossaryentry{dl}{name=DL, description={Deep Learning},first={Deep Learning (DL)}}
% \newglossaryentry{dnn}{name=DNNs, description={Deep Neural Networks},first={Deep Neural Networks (DNNs)}}
%\newglossaryentry{dgm}{name=DGMs, description={Deep Generative Models},first={Deep Generative Models (DGMs)}}
\newglossaryentry{dgm}{name=DGM, description={Deep Generative Model},first={Deep Generative Model (DGM)}}
%e
\newglossaryentry{ee}{name=EE, description={End Effectors},first={End Effectors (EE)}}
%f
\newglossaryentry{fpr}{name=FPR, description={False Positive Rate},first={False Positive Rate (FPR)}}
%g
\newglossaryentry{gan}{name=GAN, description={Generative Adversarial Network},first={Generative Adversarial Networks (GAN)}}
\newglossaryentry{cgan}{name=cGAN, description={Conditional Generative Adversarial Network},first={Conditional Generative Adversarial Network (cGAN)}}
\newglossaryentry{nf}{name=NFs, description={Normalizing Flows},first={Normalizing Flows (NFs)}}
\newglossaryentry{cnf}{name=cNF, description={Conditional Normalizing Flow},first={Conditional Normalizing Flows (cNF)}}
% \newglossaryentry{gm}{name=MoG, description={Mixture of Gaussians Distributions},first={Mixture of Gaussians (MoG)}}
% \newglossaryentry{gmm}{name=GMMs, description={Gaussian Mixture Models},first={Gaussian Mixture Models (GMMs)}}
%i
\newglossaryentry{iid}{name=\textit{iid}, description={independent and identically distributed},first={independent and identically distributed (\textit{iid})}}
% \newglossaryentry{ib}{name=IB, description={Information Bottleneck},first={Information Bottleneck (IB)}}
% \newglossaryentry{id}{name=ID, description={In-Distribution},first={In-Distribution (ID)}}
%k
\newglossaryentry{kld}{name=KLD, description={Kullaback-Leibler Divergence},first={Kullaback-Leibler Divergence (KLD)}}
%l
\newglossaryentry{ll}{name=LL, description={Log-Likelihood},first={Log-Likelihood (LL)}}
\newglossaryentry{lars}{name=LARS, description={Learned accept/reject sampling},first={Learned accept/reject sampling (LARS)}}
\newglossaryentry{llm}{name=LLM, description={Large Language Models},first={Large Language Models (LLM)}}

%m
\newglossaryentry{mllm}{name=MLLM, description={Multimodal Large Language Models},first={Multimodal Large Language Models (MLLM)}}

\newglossaryentry{knn}{name=KNN, description={k-Nearest Neighbors},first={k-Nearest Neighbors (KNN)}}

\newglossaryentry{mle}{name=MLE, description={Maximum Likelihood Estimation},first={Maximum Likelihood Estimation (MLE)}}
% \newglossaryentry{mi}{name=MI, description={Mutual Information},first={Mutual Information (MI)}}
% \newglossaryentry{ml}{name=ML, description={Machine Learning},first={Machine Learning (ML)}}
% \newglossaryentry{magd}{name=MAGD, description={Mean Absolute Grasp Deviation},first={Mean Absolute Grasp Deviation (MAGD)}}
\newglossaryentry{mlp}{name=MLP, description={Multi-Layer Perceptron},first={Multi-Layer Perceptron (MLP)}}
%o
% \newglossaryentry{ood}{name=OOD, description={Out-of-Distribution},first={Out-of-Distribution (OOD)}}
%r
% \newglossaryentry{rsb}{name=RSB, description={Resampled Base Distributions},first={Resampled Base Distributions (RSB)}}
%s
% \newglossaryentry{sgvb}{name=SGVB, description={Stochastic Gradient Variational Bayes},first={Stochastic Gradient Variational Bayes (SGVB) }}
%t
% \newglossaryentry{tpr}{name=TPR, description={True Positive Rate},first={True Positive Rate(TPR)}}
%v
\newglossaryentry{vae}{name=VAE, description={Variational Autoencoder},first={Variational Autoencoder (VAE)}}
\newglossaryentry{vi}{name=VI, description={Variational Inference},first={Variational Inference (VI)}}
\newglossaryentry{vlm}{name=VLM, description={Vision Language Models},first={Vision Language Models (VLM)}}
\newglossaryentry{magd}{name=MAGD, description={Mean Absolute Grasp Deviation},first={Mean Absolute Grasp Deviation (MAGD)}}

%%%%%%%%%%%%%%%%%%%%%%%%%%%%%%%%%%%%%%%%%%%%%%%%%%%%%%%%%%%%%%%%%%%%%%%%%%%%%%%%
\begin{abstract}
% original version
% Few-shot dexterous manipulation towards novel scenes is a challenging but essential problem as real-world dexterous hand data are expensive. We present Language-ENhanced Sparse Distilled Feature Field (LensDFF), Language-enhanced Sparse Feature Distillation to efficiently distill 2D view-inconsistent features onto 3D points with language enhancement in a zero-shot manner. Together with grasp primitives being injected in few-shot demonstrations, the proposed framework achieves efficient and consistent feature distillation and stable grasping with high dexterity. Moreover, a physically plausible real2sim grasp evaluation pipeline is proposed for efficient grasp evaluation. Through extensive simulation experiments based on the real2sim pipeline and real-world experiments, our pipeline achieves decent grasping performance, outperforming state-of-the-art methods.

% modified by JX
Learning dexterous manipulation from few-shot demonstrations is a significant yet challenging problem for advanced, human-like robotic systems. 
Dense distilled feature fields have addressed this challenge by distilling rich semantic features from 2D visual foundation models into the 3D domain. However, their reliance on neural rendering models such as Neural Radiance Fields (NeRF) or Gaussian Splatting results in high computational costs.
In contrast, previous approaches based on sparse feature fields either suffer from inefficiencies due to multi-view dependencies and extensive training or lack sufficient grasp dexterity.
To overcome these limitations, we propose Language-ENhanced Sparse Distilled Feature Field (LensDFF), which efficiently distills view-consistent 2D features onto 3D points using our novel language-enhanced feature fusion strategy, thereby enabling single-view few-shot generalization. 
Based on LensDFF, we further introduce a few-shot dexterous manipulation framework that integrates grasp primitives into the demonstrations to generate stable and highly dexterous grasps. 
Moreover, we present a real2sim grasp evaluation pipeline for efficient grasp assessment and hyperparameter tuning. 
Through extensive simulation experiments based on the real2sim pipeline and real-world experiments, our approach achieves competitive grasping performance, outperforming state-of-the-art approaches.

\end{abstract}

%%%%%%%%%%%%%%%%%%%%%%%%%%%%%%%%%%%%%%%%%%%%%%%%%%%%%%%%%%%%%%%%%%%%%%%%%%%%%%%%

\newcommand{\TODO}[1]{\textcolor{red}{TODO:#1}}
\newcommand{\hand}{DLR-HIT Hand II}

\section{Introduction}

% why manipulation is important
% Robotic manipulation is an essential technique for advanced robotic applications.
% why use hand
Recently, dexterous grasping has garnered sigfinicant attention as it pushes the boundaries of robotic manipulation towards human-like proficiency.
%enabling robots to interact with objects in unstructured environments with higher manipulability and dexterity.
% why few-shot learning is important for hand
Data-driven approaches for high-DoF robotic hands~\cite{ffhnet,zhang2024dexgraspnet20,feng2024ffhflowflow,weng2024dexdiffuser,liu2020deepdifferentiablegrasp} often require large-scale synthetic dataset~\cite{wang2023dexgraspnet} for training, which inevitably introduce a sim2real gap. On the other hand, while real-world data is more realistic, it is prohibitively expensive to collect. Therefore, developing efficient few-shot learning techniques to endow dexterous robotic systems with generalizable manipulation capabilities is both essential and challenging.
% % inspire from human
% We, as human, demonstrate superior abilities to learn new skills from observed demonstrations~\cite{lake2015concept,lake2023human}. 
% recent techniques

Recent advancements in vision-language models (VLMs) such as CLIP~\cite{radford2021clip}, SAM~\cite{kirillov2023sam} and Dino~\cite{oquab2024dinov2} have opened new possibilities for enabling robots to perform manipulation tasks with minimal training data~\cite{shen2023f3rm,rashid2023lerftogo,wang2024sparsedff,zheng2024gaussiangrasper,ji2024graspsplats}.
Since effective interaction with the environment requires accurate 3D information, a straightforward approach is to extract 2D semantic features from these vision models~\cite{radford2021clip,oquab2024dinov2} and fuse them into a 3D point representation. However, this fusion strategy often suffers from semantic inconsistency across views, as the 2D features are not inherently aligned across multiple viewpoints.
To address this challenge, distilled feature fields (DFF)~\cite{kobayashi2022decomposingnerf,kerr2023lerf} has proposed reconstructing 3D feature fields from 2D images with neural implicit representations. Building on this idea, several studies~\cite{shen2023f3rm,rashid2023lerftogo,zheng2024gaussiangrasper,ji2024graspsplats} have demonstrated promising performance in both scene understanding and language-guided manipulation. For instance, some methods~\cite{shen2023f3rm,kerr2023lerf} rely on dense view acquisition for training and scene construction (e.g. 50 views in F3RM~\cite{shen2023f3rm}), whereas others~\cite{zheng2024gaussiangrasper,ji2024graspsplats} improve efficiency by reducing the viewpoints to just 5. Nevertheless, most of these approaches have focused primarily on parallel-jaw grippers and require additional training effort. In contrast, only a few works~\cite {wang2024sparsedff,wang2024neuralattentionfield} have explored dexterous hands. However, these methods depend on feature alignment networks to reconcile inconsistent features, and the full potential of hand dexterity remains underexplored.

% Our approach
In this work, we propose LensDFF and develop an efficient few-shot dexterous manipulation framework that enables grasping of novel objects from a single view with high dexterity using grasp primitives~\cite{ciocarlie2009hand}.
Concretely, our approach introduces a novel and efficient way of utilizing language features to align view-inconsistent features without requiring any additional training or fine-tuning.
This idea of applying language features to tackle the view-inconsistency issue is motivated by the observation that language features possess a more steady semantic understanding because they are less sensitive to variations in lighting and color, compared to their vision counterparts.
Moreover, insights from neuroscience and psychology~\cite{Arbib2008FromGT,IVERSONJANAM2010Dlia} suggest a strong correlation between human motor skill learning, such as grasping, and language acquisition.

LensDFF employs language-enhanced feature alignment to adaptively project vision features from sparse views onto language features extracted from CLIP~\cite{radford2021clip}, thereby mitigating the challenge of view inconsistency and eliminating the need for additional training or fine-tuning. Consequently, our method enables dexterous robotic hands to execute robust grasps while maintaining high adaptability across novel objects and scenarios. Our contributions can be summarized:
\begin{figure*}[htbp]
    \vspace{5pt}
    \centering
    \includegraphics[width=0.85\textwidth]{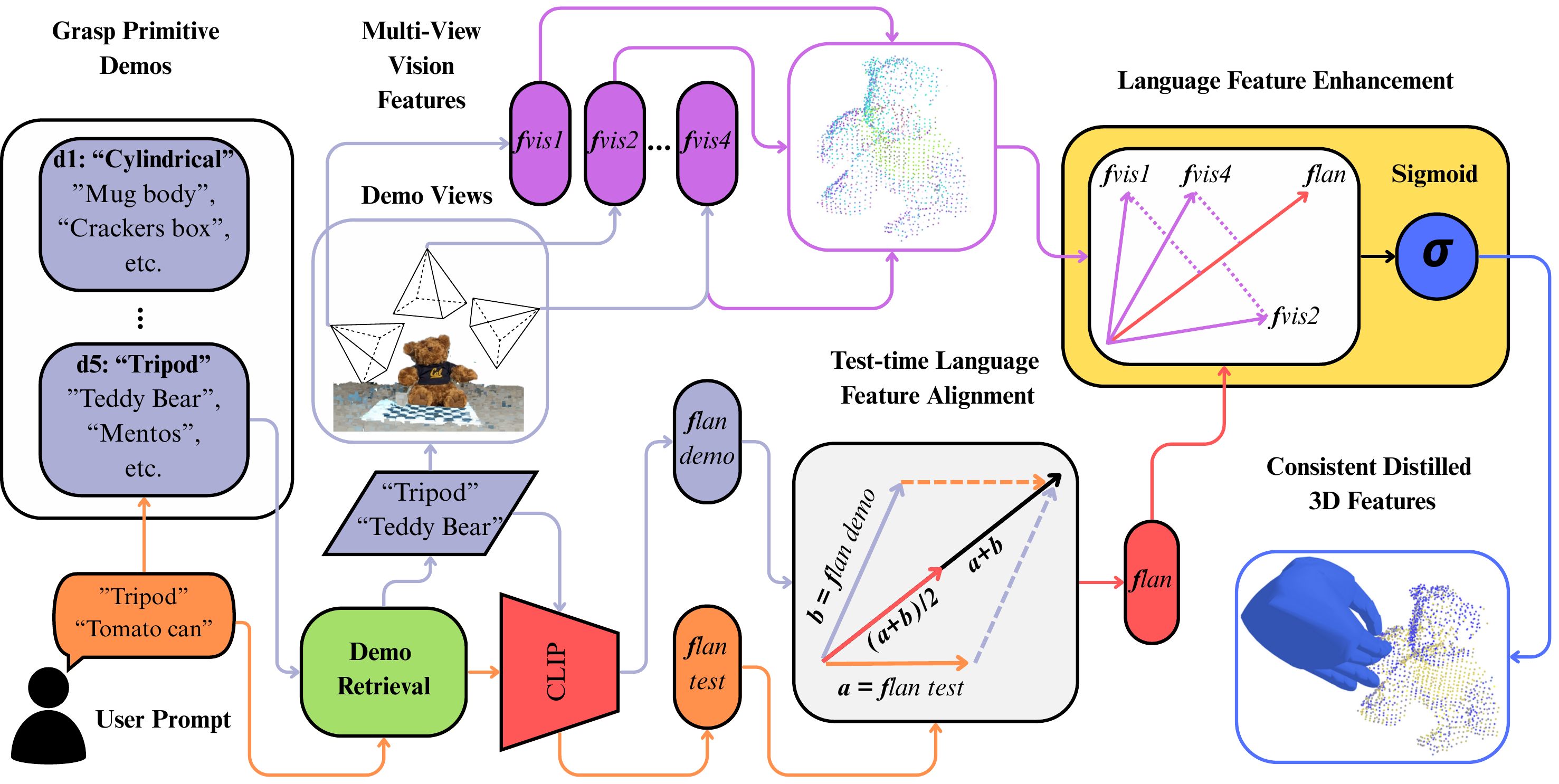}
    \caption{LensDFF demo data pipeline. Given a user prompt including the object name and grasp primitive, a closest demo is retrieved where their demo prompt features $\mathbf{f}_{\text{lan}}^{\text{demo}}$ are compared with test prompt features $\mathbf{f}_{\text{lan}}^{\text{test}}$ for \textbf{test-time language feature alignment}. The resulting language feature is then used for \text{language feature enhancement}, aligning vision features $\mathbf{f}_{\text{vis}}$ from multiple demo viewpoints to generate consistent \textbf{distilled 3D features}. }
    \label{fig:feature-distill}
    \vspace{-5pt}
\end{figure*}
% Our contributions:
\begin{itemize}
\item We propose Language-enhanced Sparse Feature Distillation (LensDFF), a novel vision feature alignment strategy that leverages language features to enable robotic manipulation with no extra training or finetuning. 
\item We propose an efficient few-shot grasp-primitive-based dexterous grasping framework built upon LensDFF, achieving stable and highly dexterous grasping of unseen objects from a single view.
\item A novel real2sim grasp evaluation pipeline for general few-shot dexterous grasping.  
\item Extensive simulation and real-world experiments validate the effectiveness of our proposed methods, outperforming state-of-the-art approaches.
\end{itemize}

%===============================================================================
% To be reviewed
\section{Related Work}

\subsection{Feature Field for Manipulation}
% this one is from chatgpt
Recent advances in neural representations like NeRF~\cite{mildenhall2020nerf}, Gaussian Splatting~\cite{kerbl20233dgaussiansplatt}, have not only revolutionized view synthesis but have also found applications in robotics~\cite{simeonov2021neuraldescriptorfield,dai2023graspnerf}.
Furthermore, researchers have demonstrated that combining feature distillation from 2D foundation models with neural rendering can yield high-quality representations that enable robotic manipulation~\cite{shen2023f3rm,rashid2023lerftogo,ji2024graspsplats,zheng2024gaussiangrasper}.

Specifically, the works F3RM~\cite{shen2023f3rm} and LERF-TOGO~\cite{rashid2023lerftogo} distill features from foundation models such as CLIP~\cite{radford2021clip}, SAM~\cite{kirillov2023sam}, and Dino~\cite{oquab2024dinov2} into 3D scenes to enable language-guided grasping. However, these approaches require the collection of dense viewpoints and additional training for each scene. F3RM~\cite{shen2023f3rm}, for example, requires about 1m 40s for data collection and 3 minutes for feature distillation during our replication.
To reduce the time, some works~\cite{ji2024graspsplats,zheng2024gaussiangrasper}
employ more efficient feature distillation methods based on 3D Gaussian representations to speed up the feature field reconstruction to about 1 minute. 

Nevertheless, most aforementioned works focus on robotic tasks involving parallel-jaw grippers, which inherently limit task complexity and overall manipulability. Only a few studies have addressed the challenge of few-shot dexterous grasping using DFF~\cite{wang2024sparsedff, wang2024neuralattentionfield}. These approaches typically propose a feature alignment network to align features from sparse views, but their frameworks struggle to efficiently handle high-dimension language features. In contrast, our framework achieves 3D point feature alignment using language features, eliminating the need for extra training or finetuning. Moreover, we incorporate grasp primitives in our few-shot demonstrations to enhance dexterity and manipulability.

\subsection{Dexterous Grasping}
Analytical approaches rely on the hand and the object geometries to generate grasp samples, using hand-crafted geometric constraints, heuristics, and point cloud features~\cite{Lei2017cshape, lu2017grasp, Lu2019Reconstruct}. 
Learning-based approaches for dexterous grasping can be broadly divided into generative-model-based and regression-based approaches.
Generative-model-based approaches~\cite{ffhnet,feng2024ffhflowflow,weng2024dexdiffuser,feng2024dexgangrasp,zhang2024dexgraspnet20,wei2024funcgrasp} integrate grasp generation and optimization, but they often require substaintial effort to balance grasp stability, diversity, and runtime. Meanwhile, regression-based approaches~\cite{Li2022hgcnet,liu2020deepdifferentiablegrasp,chen2022isagrasp} directly predict grasp poses, neglecting the inherent multimodality of grasp distributions. 
Moreover, all of these methods typically depend on training a dedicated grasping model using large synthetic datasets which inevitably introduces a sim2real gap. For instance, although a few studies~\cite{wei2024funcgrasp,chen2022isagrasp} employ few-shot demonstrations, they still generate extensive synthetic datasets for the training. 

Only a handful of works~\cite {jeremy2016oneshot, wang2024sparsedff, wang2024neuralattentionfield} address the challenge of few-shot dexterous grasping. Among these, the approach in~\cite{jeremy2016oneshot}, which combines few-shot or one-shot methods with grasp types, replies on hand-crafted geometric features from the test objects and carefully designed modeling of contacts and hand configurations. In contrast, our method leverages vision and language features distilled onto 3D point clouds to enable language-guided manipulation with more efficient optimization.

% \subsection{Grasp Primitives}

% \cite{smith2020grasping}
% \cite{ciocarlie2009hand} -> eigengrasp

% "Grasp Type Revisited: A Modern Perspective on A Classical Feature for Vision"

% "An adaptive planning framework for dexterous robotic grasping with grasp type detection"
% A neural network is proposed to predict the grasp type given visual inputs, to guide the grasp generation. 

% \subsection{Active Learning}
% Few-Shot Continual Active Learning by a Robot
%  learn new objects over time with minimal supervision while avoiding catastrophic forgetting
 
%  Gaussian Mixture Model-Based Continual Learning (GBCL)
 
%  actively selects the most uncertain objects to be labeled
% Predictive Entropy: Measures how uncertain the system is about an object's classification.
% Viewpoint Consistency: Evaluates prediction consistency across different angles of an object.

%===============================================================================
\section{Method}

\begin{figure*}[t!]
\vspace{5pt}
    \centering
    \includegraphics[width=0.85\textwidth]{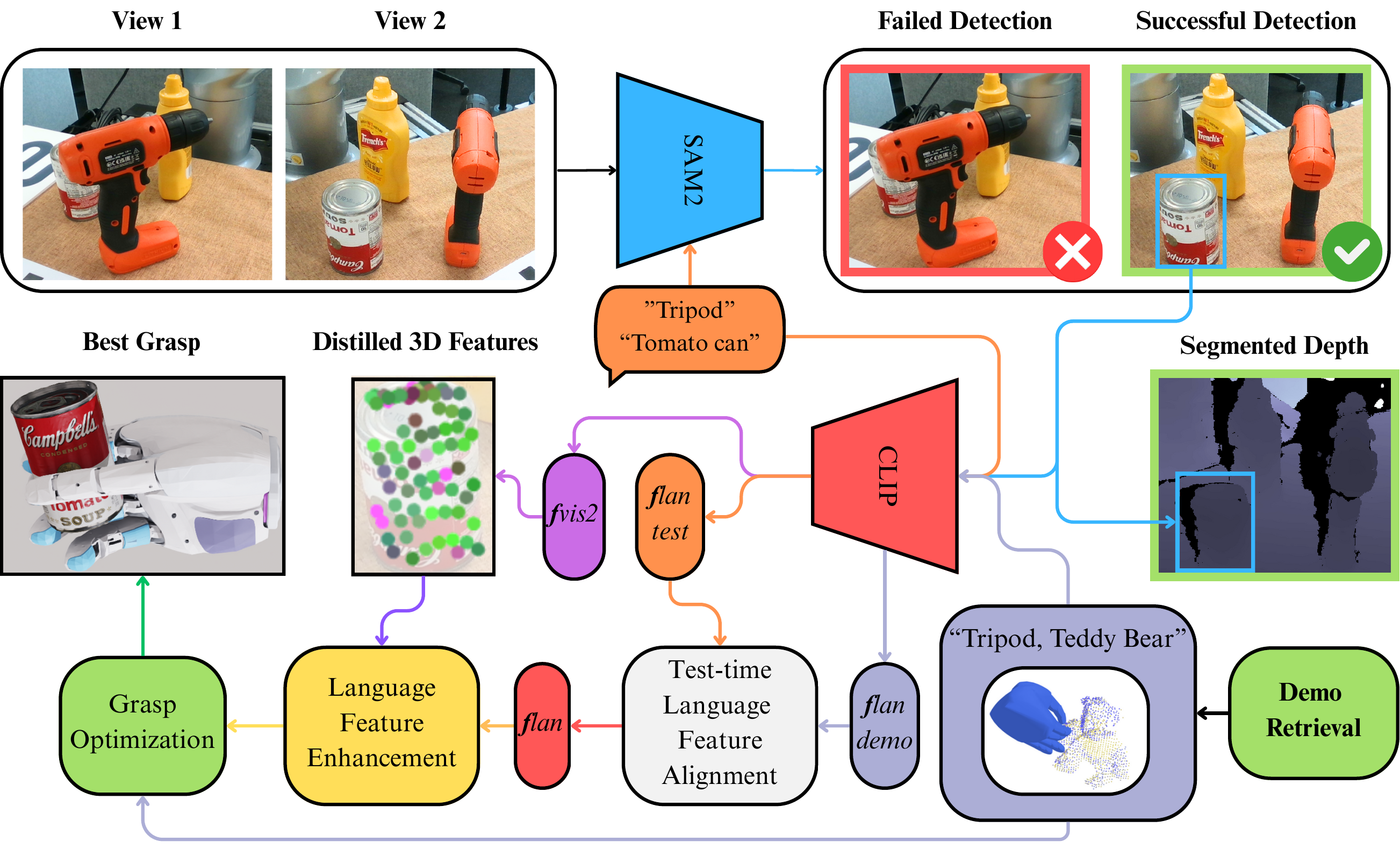}
    \vspace{5pt}
    \caption{LensDFF test data pipeline. Our approach applies SAM2\cite{ravi2024sam2} to a single RGB image to detect the target object. A second view is selected if the object is not visible. The same \textbf{test-time language feature alignment} and \textbf{language feature enhancement}  as in the demo data pipeline are applied. The main difference is that only vision features from one view are projected. Finally, the \textbf{3D distilled features} from both the demo and test data are utilized for grasp optimization.}
    \label{fig:feature-distill-2}
    \vspace{-8pt}
\end{figure*}
\subsection{Problem Formulation}
We assume a robot arm equipped with a dexterous \hand~ and a single eye-in-hand calibrated RGBD camera.  
A grasp $\mathbf{g} \in \mathbb{R}^{24}$ is represented by the 15-DOF hand joint configuration $\boldsymbol{j} \in \mathbb{R}^{15}$ and the 6D pose $(\mathbf{R},\mathbf{t}) \in SE(3)$ of the palm, where the rotation $\mathbf{R}$ is expressed using a continuous 6D representation~\cite{zhou2019continuity}.

During the demo collection, the robot captures the demo object from a \textbf{sparse} set of viewpoints (e.g., 4). 
Each demo consists of a ground truth grasp pose $\mathbf{g_{gt}}$ which is teleoperated by the human expert, along with a text prompt describing the demo object $p$, point clouds $X_i$ and color images $I_i$ from each viewpoint $i=[1,2,3,4]$. 
During testing, the robot captures an observation from a \textbf{single} viewpoint. If the target object is unrecognized due to clutter or occlusions, a next viewpoint is selected. The goal is to pick up the target object based on a language prompt that describes the object and a specified grasp primitive.

% Apply CLIP for each view
% Given several camera observations from a demonstration, our approach first extracts features $F$ from every 2D image $I$. % onto 3D point clouds $X$ with pixel-point correspondences.

\subsection{3D Language-enhanced Sparse Feature Distillation}
% CLIP background (This has to be deeper understood)
\paragraph{3D Sparse-View Feature Distillation} After collecting demonstrations, we first obtain the bounding box (bbox) and segmentation masks via SAM2~\cite{ravi2024sam2}, given a demonstration prompt $p$. 
The cropped image from the bbox is then processed by CLIP to extract vision features, while the segmentation mask enables the extraction of pixel-level vision features and their corresponding 3D points.
Consistent and meaningful 3D features are essential for effective matching with the test scene in robot manipulation.
VLMs such as CLIP~\cite{radford2021clip} have shown strong capabilities in extracting vision-language aligned features. 
However, these features are typically aligned as a whole instead of pixel level, where a fine-grained alignment is missing~\cite{zhong2021regionclip}.

% cannot simple merge
\paragraph{Language Feature Enhancement} Thus, simply merging point clouds with projected vision features results in view-inconsistent 3object surface 3D features due to a lack of 3D awareness in the 2D vision foundation model. 
To address this issue, rather than training an additional alignment network purely on vision features, as in~\cite{wang2024sparsedff}, we propose an efficient language-enhanced feature distillation strategy that aligns features cross views which requires~\textbf{no extra training or fine-tuning}.
The key intuition behind employing language features is that, while CLIP~\cite{radford2021clip} vision features exhibit view inconsistency, language features remain more stable and provide consistent semantic representations across views.
Given a demonstration prompt $p$ describing the object, we project CLIP vision features for each point $x_i$ onto the corresponding language feature $\mathbf{f}_{lan}$ ensuring better feature. 
Empirically, we find that this approach effectively achieves a good balance in incorporating both feature types, i.e., preserving the magnitude of the multi-view vision features while aligning them with the direction of the language feature (Fig. \ref{fig:feature-distill}). 
\begin{equation}
\label{eq:lan_enhancement}
\mathbf{f}_i^{\text{aligned}}
= \sigma\!\Bigl(
   \frac{\langle \mathbf{f}_{\text{vis}}(x_i),\, \mathbf{f}_{\text{lan}} \rangle}
        {\|\mathbf{f}_{\text{lan}}\|^2}
\Bigr)\, \mathbf{f}_{\text{lan}}.
\end{equation}
The projected features are sent to the sigmoid activation function $\sigma$ for better normalization and interpretability.
More details in Fig.\ref{fig:feature-distill}.
% \textbf{Single View Pixel-aligned Features}
% Inspired by~\cite{jata2023conceptfusion}, the pixel-aligned features for the target objects detected with GroundedSAM~\cite{ren2024groundedsam}, 
% is computed with a weighted combination of global features $\mathbf{f}_{\text{vis}}^{G}$ and its local features $\mathbf{f}_{\text{vis}}^{L}$. The weight is obtained through the consine similarity between these two features. 

% \[
% \text{cos\_sim}(\mathbf{f}_{\text{vis}}^{G}, \mathbf{f}_{\text{vis}}^{L}) 
% \right\rangle 
% \;=\; \frac{\left(\mathbf{f}_{\text{vis}}^{L}\right)^{\mathsf{T}} \,\mathbf{f}_{\text{vis}}^{G}}
%        {\|\mathbf{f}_{\text{vis}}^{L}\|\;\|\mathbf{f}_{\text{vis}}^{G}\|\;+\;\epsilon}.
% \]

% \[ 
% w_i = \text{cos\_sim}(\mathbf{f}_{\text{vis}}^{G}, \mathbf{f}_{\text{vis}}^{L}) 
% \]

% where $\epsilon$ is used as a small value, e.g., $1e-8$ to avoid division by zero. Using the cosine similarity as the weight, we can compute the pixel-aligned vision local features:

% \[
% \mathbf{f}_{\text{vis}}^{\text{aligned}}
% \;=\;
% w_i\,\mathbf{f}_{\text{vis}}^{G}
% \;+\;
% \bigl(1 - w_i\bigr)\,\mathbf{f}_{\text{vis}}^{L}.
% \]

% \begin{figure}
% \vspace{5pt}
%     \centering
%     \includegraphics[width=0.4\textwidth]{materials/3_method/feature_aling.pdf}
%     \vspace{5pt}
%     \caption{Feature alignment}
%     \label{fig:robot_setup}
% \end{figure}

\paragraph{Test-time Language Feature Alignment}

At test-time, the inferred grasp may fail if the demo object prompt features $\mathbf{f}_{\text{lan}}^{\text{demo}}$ differ significantly from the test object prompt features $\mathbf{f}_{\text{lan}}^{\text{test}}$. To address this, we propose an adaptive language alignment strategy during inference. 
This strategy computes the cosine similarity $s$ between these two language features. 
If $s$ exceeds a threshold, indicating that the test object prompt is sufficiently similar to the demo prompt, we directly use $\mathbf{f}_{\text{lan}}^{\text{demo}}$. 
Otherwise, we fuse both language features to account for their difference between them, ensuring a smother generalization to novel objects, shown in Fig.\ref{fig:feature-distill}.

\begin{equation}
\mathbf{f}_{\text{lan}} =
\begin{cases}
    \mathbf{f}_{\text{lan}}^{\text{demo}}, & \text{if } s \geq \tau, \\
    (\mathbf{f}_{\text{lan}}^{\text{demo}} + \mathbf{f}_{\text{lan}}^{\text{test}}) / 2, 
    & \text{otherwise}.
\end{cases}
\end{equation}, where
\begin{equation}
s = \frac{\mathbf{f}_{\text{lan}}^{\text{demo}} \cdot \mathbf{f}_{\text{lan}}^{\text{test}}}
    {\|\mathbf{f}_{\text{lan}}^{\text{demo}}\| \|\mathbf{f}_{\text{lan}}^{\text{test}}\|}.
\end{equation}
We empirically set $\tau = 0.63$, detailed in Section.~\ref{sec:real2sim}.
The same strategy applies to test objects as well with the same threshold determining if projection on test object langauge feature $\mathbf{f}_{\text{lan}}^{\text{test}}$ or the fused feature from both.

\paragraph{Grasp Representation}
% Unlike previous works of reconstructing a continuous features field using NeRF~\cite{mildenhall2020nerf} or Gaussian Splating\cite{kerbl20233dgaussiansplatt} where the grasp features can be computed by locating the grasp in 3D feature field.  
After distilling features into 3D space, similarly like~\cite{wang2024sparsedff}, the grasp features $\mathbf{f}_{\text{grasp}}$ is computed by identifying nearby 3D points $\mathbf{x}_i$ corresponding to $N$ sampled hand surface points $q$ and aggregating their aligned features. The aggregation is weighted as $w_i$ by the inverse of the L2 distance, ensuring a smooth and spatially aware feature representation.

\begin{equation}
    \mathbf{f}_{\text{grasp}} =
    \sum_{i=1}^{N} w_i \mathbf{f}_i^{\text{aligned}} 
    % \text{where} \quad  
    % w_i =
    % \frac{\displaystyle \frac{1}{\|\mathbf{q} - \mathbf{x}_i\|^2}}
    % {\displaystyle \sum_{j=1}^{N} \frac{1}{\|\mathbf{q} - \mathbf{x}_j\|^2}}.
\end{equation}

This framework is applied to both demo and test RGBD images to extract multi-view consistent, pixel-level 3D features, which are then utilized for grasp optimization.

% \subsection{Hierarchical Demo Retrieval}
% During test, we need to retrieve the proper grasp primitive and the optimal demonstration before grasp optimization.

% For an efficient demo retrieval, a hierarchical structure of grasp primitives and grasp demonstrations are depicted in {fig}.

% The retrieval first find the proper grasp primitive by computing the cosine similarity between the task prompt $p_t$ and a set of graps primitive prompts $p_k$, where $k \;\in\; \text{[hook, cylindrical, pinch, tripod and lumbrical]}$.

% \begin{align*}
% \text{cos\_sim}(\mathbf{p_t}, \mathbf{p}_k) 
% &= \frac{\mathbf{t} \cdot \mathbf{p}_k}{\|\mathbf{t}\|\;\|\mathbf{p}_k\|} \\ 
% k^* &= \arg\max_{k}\; \text{cos\_sim}(\mathbf{t}, \mathbf{p}_k)
% \end{align*}

\subsection{Grasp Demonstration with Primitives}

\begin{figure*}[htbp]
    \vspace{6pt}
    \centering
    \includegraphics[width=0.86\textwidth]{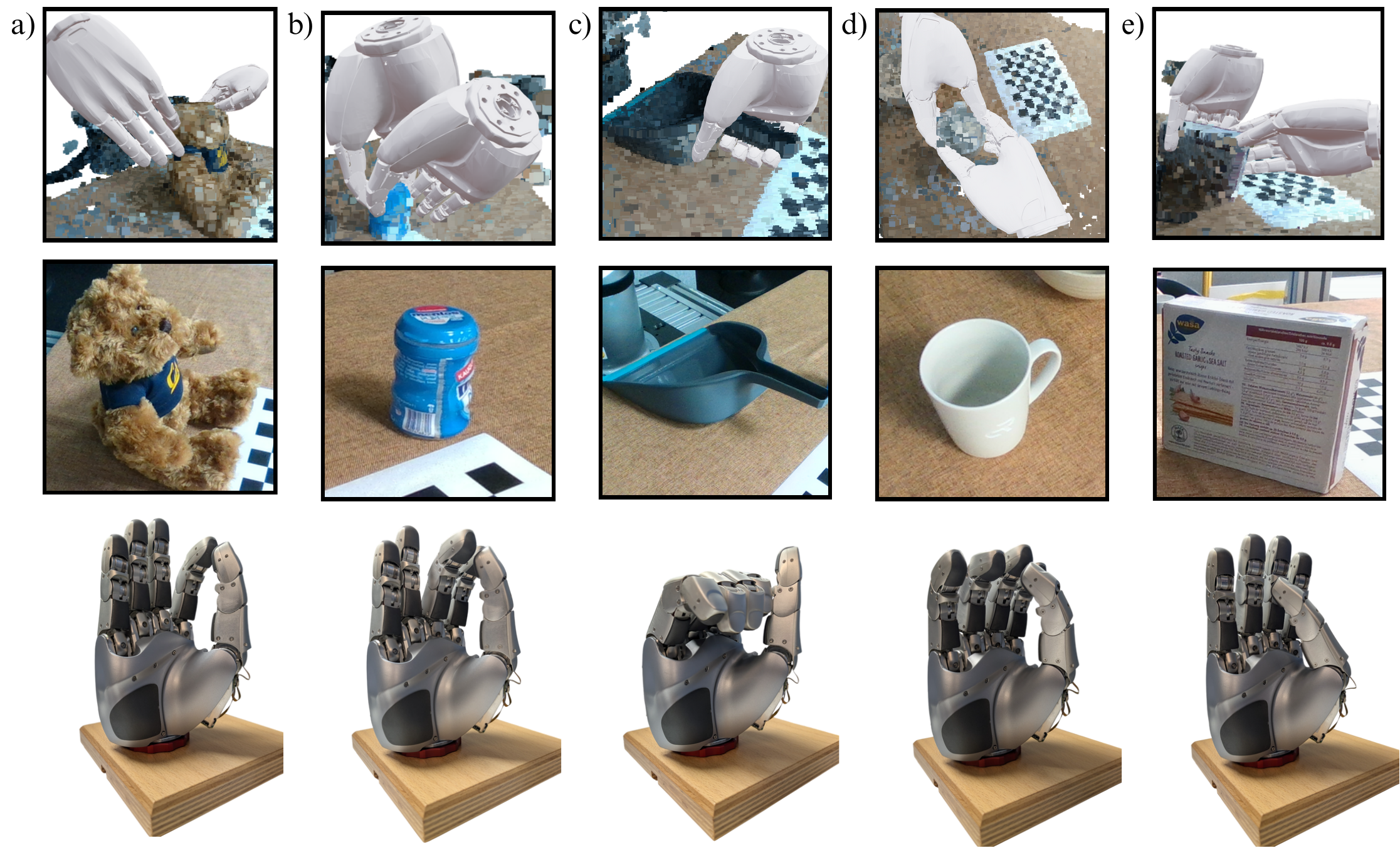}
    \vspace{-5pt}
    \caption{Demo Grasps with Diverse Grasp Primitives.
    This figure illustrates the versatility of our collected demos using different grasp primitives across a range of objects.
    (a) \textbf{Pinch grasp}: The robot delicately pinches the teddy bear's ear between the thumb and index finger, demonstrating precision and control for handling small or delicate objects. 
    (b) \textbf{Hook grasp}: The robot secures the handle of a dustpan using a hook grasp, forming hooks with its fingers to ensure a firm grip for lifting or carrying.
    (c) \textbf{Tripod grasp}: The Mentos gum package is grasped with a tripod grasp, where the thumb and two fingers provide stability and dexterity for precise manipulation.
    (d) \textbf{Cylindrical grasp}: The robot wraps its fingers around the white mug, forming a cylindrical grasp that ensures stability and force closure for larger objects.
    (e) \textbf{Lumbrical grasp}: The robot adopts a lumbrical grasp to hold the crackers box, with fingers are positioned parallel to the object's surface, offering a secure grip for flat or boxy objects.
    }
    \label{fig:demos}
    \vspace{-15pt}
\end{figure*}

\paragraph{Primitive Design} To equip our robotic system with dexterous manipulation capabilities using a limited set of real-world demonstrations, we employ five distinct grasp primitives: \textbf{hook, cylindrical, pinch, tripod}, and \textbf{lumbrical} grasps.
In each demonstration, a human expert selects the most suitable primitive for the task, mode details in Fig.~\ref{fig:demos}. Moreover, an object can be manipulated using multiple primitives. 
For example, when handling a cup, a \textbf{pinch grasp} is applied to the handle while a \textbf{cylindrical grasp} is used for the cup body.

\paragraph{Demo Retrieval} Since the appropriate grasp primitive for optimization is given by user, the next step is to select the most relevant demo. Each grasp primitive has multiple demo grasps available. Therefore, we follow the strategy used in F3RM~\cite{shen2023f3rm}, computing the cosine similarity between the grasp features $\mathbf{f}_{\text{grasp}}$ and the test prompt language features $\mathbf{f}_{\text{lan}}^{\text{test}}$. The demo with the closest grasp features is then selected for the optimization.

\subsection{Dexterous Grasp Inference}

\paragraph{Normal-based Grasp Initialization}

\begin{figure}[!h]
    \vspace{-2pt}
    \begin{adjustbox}{center}%{max width=1\linewidth}
    \centering
    \subfigure[Grasp frame]{
        \centering
        \includegraphics[width=0.22\columnwidth]{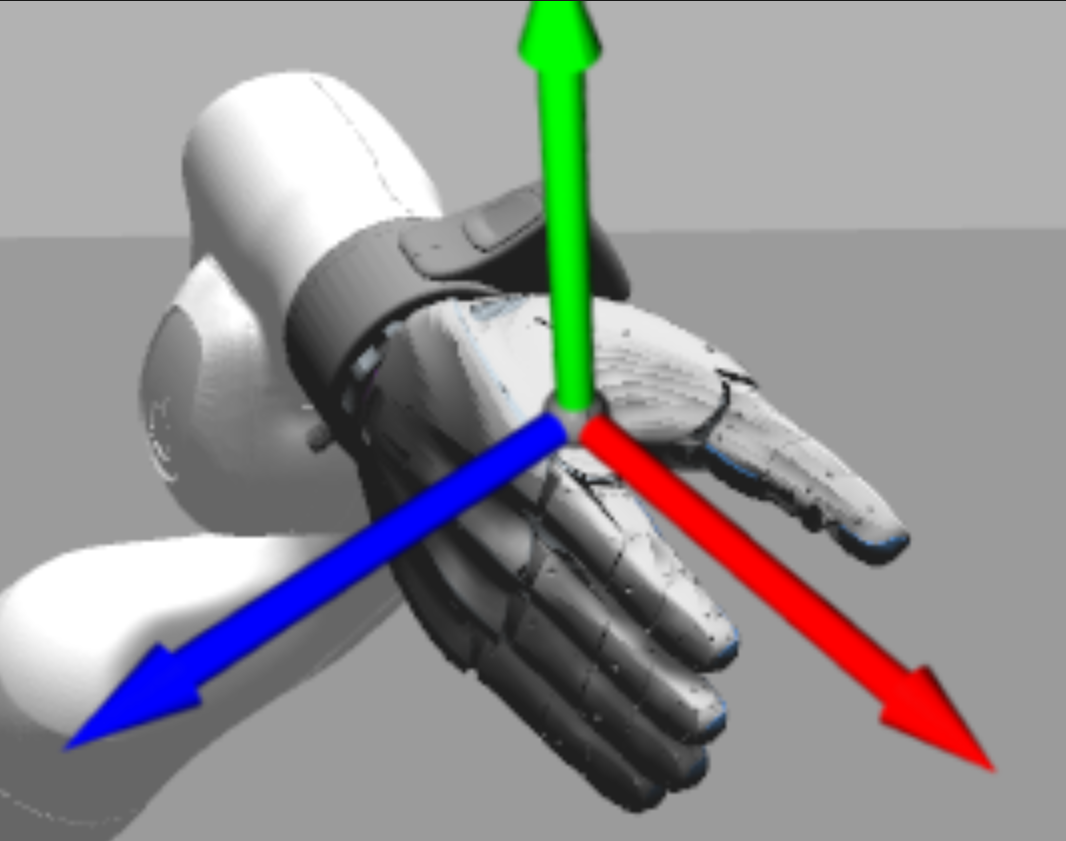}
    }
    \subfigure[Sample from point cloud normal]{
        \centering
        \includegraphics[width=0.25\columnwidth]{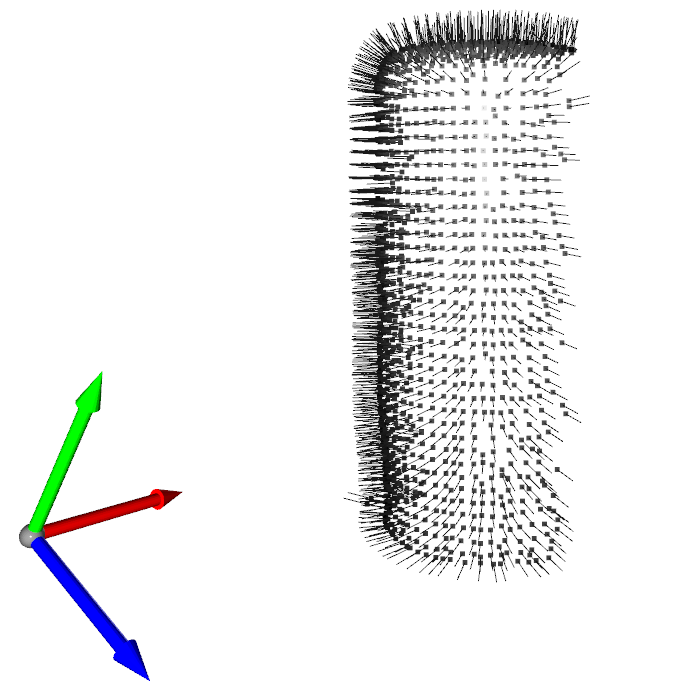}
    }
    \subfigure[Sampled poses]{
        \centering
        \includegraphics[width=0.4\columnwidth]{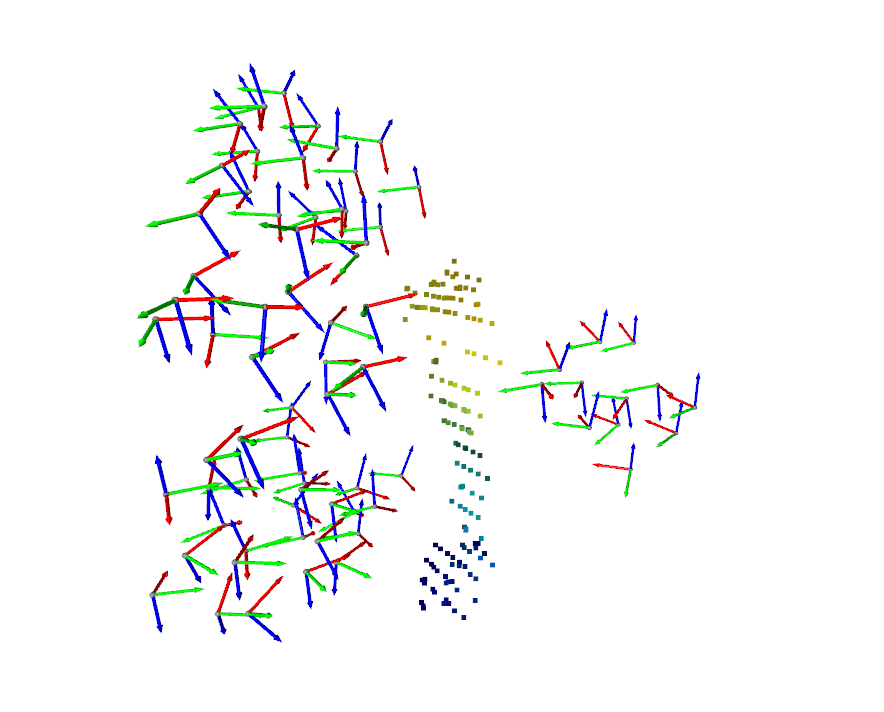}
    }
    \end{adjustbox}
    \caption{Visualization of the palm pose sampler. (c) is an example where the poses are sampled from a partial view of a drill.}
    \vspace{-3em}
    \label{fig:hand_frame}
\end{figure}

Generating diverse and well-structured grasp poses is crucial for efficient grasp optimization, especially when only single-view observations of test objects are available. The normal-based grasp sampler first determines the palm pose, followed by sequential joint configuration sampling.
A grasp frame is defined for \hand, positioned at the center of the palm and oriented between the thumb and index finger, shown in Fig.\ref{fig:hand_frame} (a). 
The x-axis of the palm pose is encouraged to align toward the objects by leveraging point cloud normals
Once the x-axis is aligned, a 3D bbox is fiited to the object point cloud, with its longest side defining the y-axis. 
To introduce variation, the sampled palm pose is perturbed with both translational and rotational noise. 
As shown in Fig.~\ref{fig:hand_frame} (c), when working with a singl-view point cloud, normal direction ambiguities can arise, leading to grasp samples being generated on both sides of the object surface.

After palm pose sampling, the joints are randomly sampled within their respective limits while adhering to constraints imposed by the chosen grasp primitives.

\paragraph{Primitive-based Grasp Optimization}
% Random sampling is stupid
We define eigengrasp~\cite{ciocarlie2009hand} for each grasp primitive to reduce the dimensionality of the grasp search space. 
For instance, in a pinch grasp, only the index and thumb are active, meaning their eigengrasp governs their motion with identical joint commands while keeping the remaining fingers static. 
Similarly, in a cylindrical grasp, all fingers close simultaneously, with their eigengrasp ensuring coordinated movement through shared joint commands. Each eigengrasp defines a mapping matrix $\mathbf{M}$, projecting the grasp pose from high-dimensional space into a low-dimensional representation.

% Instead, we build upon the idea from Ciocarlie \cite{ciocarlie2009hand} to sample in a meaningful subspace spanned by the so-called \textit{eigengrasps}.
% % Explain where the idea came from
% % Ciocarlie's work was inspired by the Neuroscience community, which showed that the joint DoFs of human hands during real-world grasping trials were largely not operating independently but coordinated. More than 80 \% of the variation in the data could be explained from the two first components of the principal component analysis (PCA).
% % What are eigengrasps
% % These components were termed eigengrasps as almost any grasp joint configuration can be synthesized as of linear combination of a few eigengrasps. 
% % Obtaining eigengrasps
% We manually define four eigengrasps $\mathbf{e_1}, \mathbf{e_2}, \mathbf{e_3}, \mathbf{e_4} \in \mathbb{R}^{15}$. The full joint configuration:
% \vspace{-1em}
% \begin{equation}
%     \mathbf{\theta} = \sum_{i=1}^4 \alpha_i \mathbf{e}_i
% \end{equation}
% % Just sample coefficient
% is obtained through sampling the coefficients $\alpha_i\in[0,1]$.
By applying eigengrasps to each grasp primitive, we obtain a simplified grasp pose $\mathbf{g_p} = \mathbf{W} \mathbf{g}$, from the original grasp pose $\mathbf{g}$, improving optimization efficiency.

The optimization objective is to minimize the difference between the grasp features $\textbf{f}_{\text{grasp}}$ extracted from the demo scene and those from the test scene.
% To avoid potential self-collision, additionally, a self-penetration energy function~\cite{wang2023dexgraspnet} is added to the optimization. 
Additionally, a normal direction constraint is enforced to ensure the final pose does not deviate excessively from the initial pose derived from the point cloud's normal direction.
\begin{equation}
E_{\text{feat}}(\mathbf{g_p}) = \left\lVert \mathbf{f}_{\text{grasp}}^{\text{demo}}
      \;-\; 
      \mathbf{f}_{\text{grasp}}^{\text{test}}(\mathbf{g_p})
      \right\rVert^2
\end{equation}

\begin{equation}
\min_{\mathbf{g_p}} E(\mathbf{g_p})
=
\underbrace{E_{\text{feat}}(\mathbf{g_p})}_{\text{feature diff.}} + 
\underbrace{\lambda_{\text{norm}} \, E_{\text{norm}}(\mathbf{g_p})}_{\text{normal restriction}}
\end{equation}

here $\lambda_{\text{normal}}$ is $1e-2$.
Every time 10 initial grasps are optimized with 300 iterations and a learning rate of $1e-2$.
% Unlike the Shadow Hand used in~\cite{wang2023dexgraspnet}, penalizing hand-object penetration is not recommended in our setup because our hand uses an impedance controller in which the target joints typically penetrate the objects in order to apply sufficient contact force. Therefore, we exclude this term from the optimization. 

%===============================================================================
\section{Experiments}

\subsection{Experimental Setup} 

\begin{figure}
    \centering
    \includegraphics[width=0.47\textwidth]{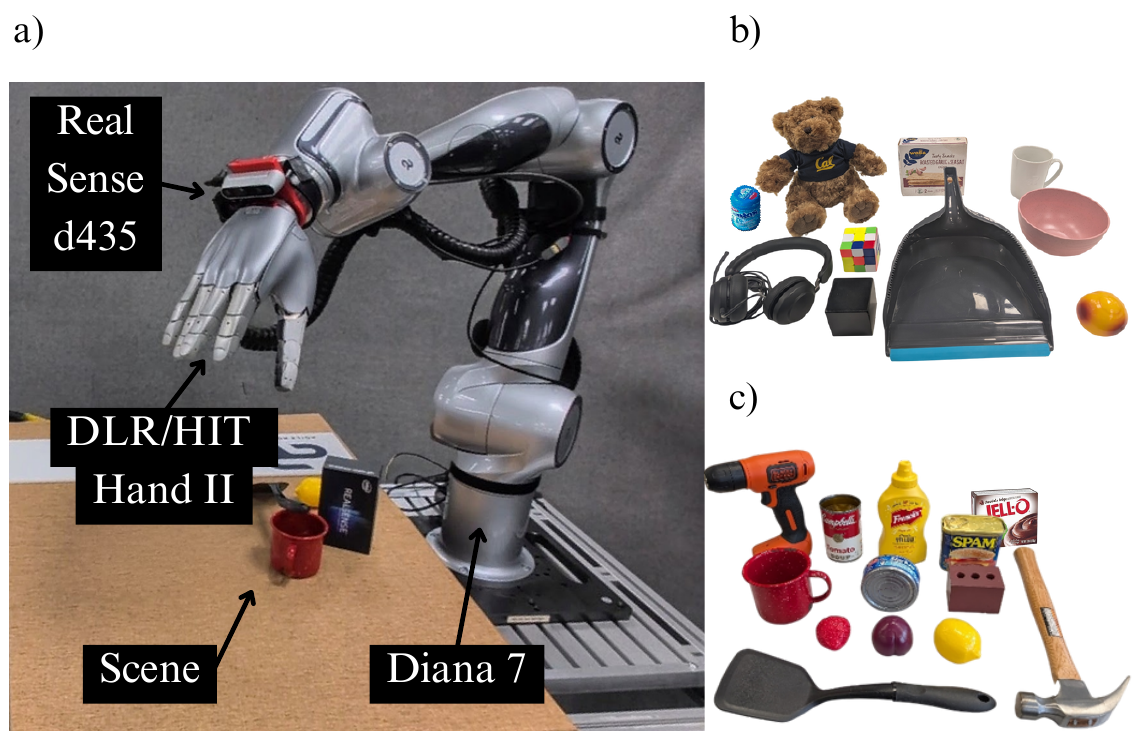}
    % \vspace{5pt}
    \caption{Real-World Experimental Setup and Objects. (a) Robot setup for real-world experiment. (b) 10 daily objects used for demo collection. (c) 12 testing YCB objects~\cite{calli2015ycb}.}
    \label{fig:robot_setup}
    \vspace{-17pt}
\end{figure}

The experimental setup consists of a Diana 7 robot arm with 7 DOFs, equipped with a \hand~ for grasping. A RealSense D435 camera mounted on the wrist is calibrated via eye-in-hand calibration. The test environment includes a table with various YCB objects~\cite{calli2015ycb} for conducting the real-world experiment and real2sim pipeline, more details shown in Fig.\ref{fig:robot_setup}.

The software framework is built on ROS2 with MoveIt for motion planning. The hand operates under joint impedance control~\cite{liu2008hithand}, ensuring stable grasp execution. Inference computations are performed on a PC equipped with an RTX A6000 GPU running Ubuntu 22.04.

We collect in total 5 demo scenes featuring 10 demo objects and 22 teleoperated demo grasps using various grasp primitive, along with user prompts describing each object.
Grasps are teleoperated using a space mouse with pre-computed grasp primitives (Fig.\ref{fig:demos}) and verified with the real hand grasp execution. 

\subsection{Real2Sim pipeline}
\label{sec:real2sim}
Since testing our pipeline in the real world is time-consuming as it would require re-scanning the scene after every grasp is executed, we propose a real-to-sim (real2sim) pipeline for the purposes of large-scale and fast grasp quality evaluation and parameter tuning. This pipeline employs SAM2~\cite{ravi2024sam2} and FoundationPose~\cite{wen2024foundationpose} to estimate the 6D poses of the objects from the YCB object set and load them in the Isaac Sim simulator~\cite{NVIDIA2023IsaacSim} with their corresponding grasps to assess success rates.

\begin{figure}[t!]
\vspace{5pt}
    \centering
    \includegraphics[width=0.48\textwidth]{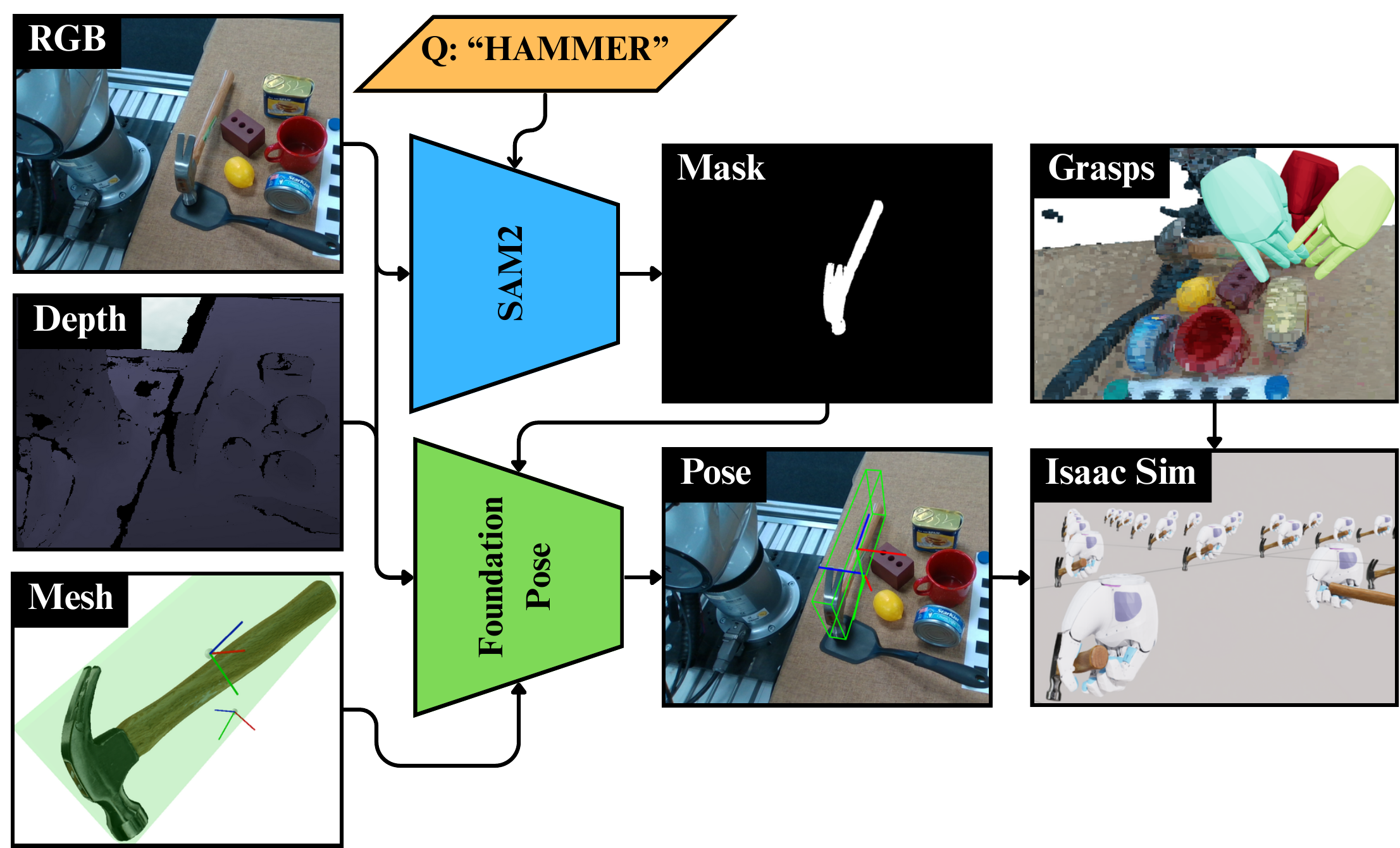}
    \caption{Real2Sim Evaluation pipeline using Isaac Sim~\cite{NVIDIA2023IsaacSim}}
    \label{fig:real2sim}
    \vspace{-5pt}
\end{figure}

% \subsubsection{Object Segmentation with SAM2}

% SAM2~\cite{ravi2024sam2} segments the target object from the scene using an RGB image and a text prompt (e.g., "Hammer"). This generates a segmentation mask that accurately isolates the object. This process follows the same approach as used in our framework.

% \subsubsection{Pose Estimation using FoundationPose}

% FoundationPose~\cite{wen2024foundationpose} estimates the 6-DOF pose of the segmented object in the camera frame by utilizing RGB images, depth data, segmentation masks, and 3D object meshes for precise pose estimation. The availability of YCB objects 3D meshes enables accurate 6D pose detection. The corresponding mesh is then spawned in the 3D viewer, allowing for manual inspection to secure the reliability of the real2sim transfer.
\subsubsection{Object Segmentation and Pose Estimation}
We use SAM2~\cite{ravi2024sam2} for segmentation using an RGB image and a text prompt (e.g., "Hammer"), followed by FoundationPose~\cite{wen2024foundationpose} to estimate the 6-DOF pose of the segmented YCB object by leveraging RGB-D images, segmentation masks from SAM2, and 3D object meshes for precise pose estimation. The estimated pose is then used to position the object in Isaac Sim, illustrated in Fig.\ref{fig:real2sim}.

\subsubsection{MultiGripperGrasp Isaac Sim Pipeline Integration}
The final step in the pipeline is to load the optimized grasp-object pairs $\mathbf{g}$ into the MultiGripperGrasp~\cite{casas2024multigripper} Isaac Sim Pipeline. The object's pose is transformed into the robot's base frame, and the grasps are executed in simulation to assess their feasibility and success rate. This pipeline supports parallel execution, allowing multiple instances of the robotic hand to grasp in different generated poses simultaneously, such as 50 instances in Fig.~\ref{fig:real2sim}. 
%The evaluation provides valuable insights into system performance and guides further refinement of hyperparameters.

% This efficient real2sim pipeline enables also test time grasp evaluation before robots execute the grasp, but because of time and effort, we leave it for future work.
\begin{figure*}[htbp]
\vspace{10pt}
    \centering
    \includegraphics[width=1\textwidth]{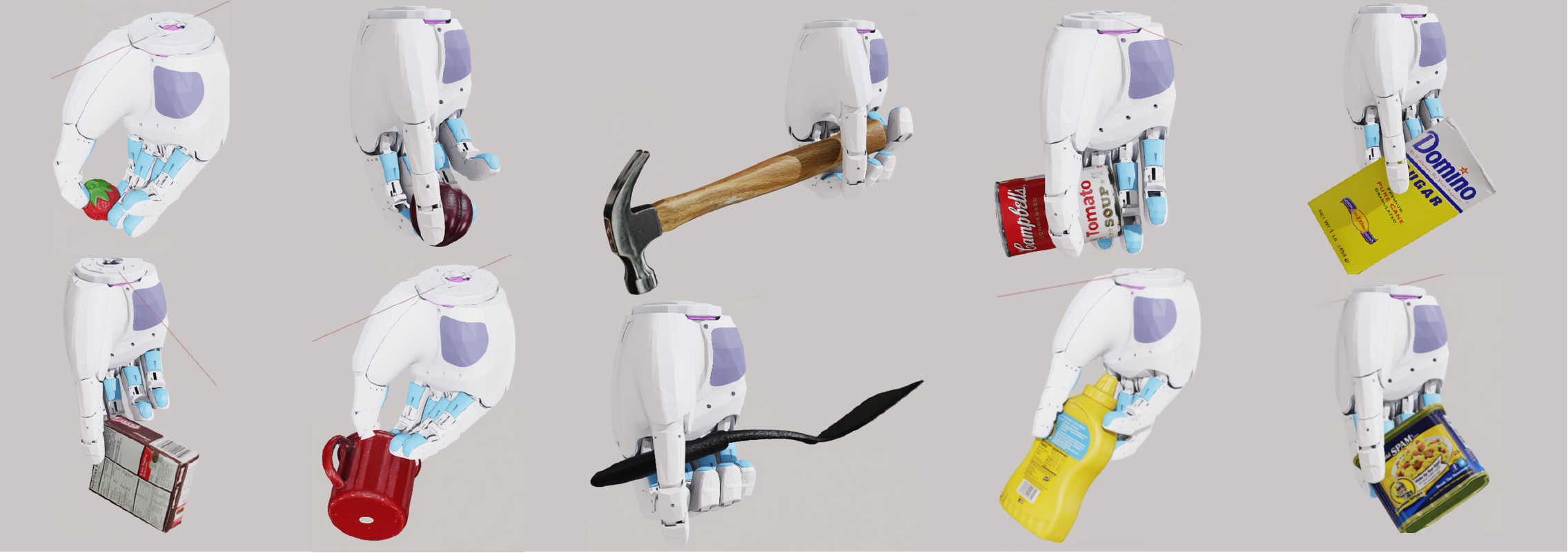}
    \caption{Isaac Sim Simulation Results of Grasping Diverse Objects. A diverse successful grasps are demonstrated, with different grasp primitives ordered in columns from left to right.
    \textbf{Pinch grasp}: The robot successfully grasps a strawberry and a Jello box, showcasing precision for manipulating small or fragile items.
    \textbf{Tripod grasp}: A plum and a red mug's lip are securely held using a tripod grasp, providing stability and dexterity for objects requiring a balance of force and precision.
    \textbf{Hook grasp}: The robot demonstrates the versatility of the hook grasp by holding a hammer and a spatula, ideal for lifting tools with handles.
    \textbf{Cylindrical grasp}: A tomato soup can and a mustard bottle are grasped securely, demonstrating stability for larger cylindrical objects.
    \textbf{Lumbrical grasp}: The robot uses a lumbrical grasp to hold a sugar box and a potted meat can, ensuring a secure grip for flat or boxy objects.}
    \vspace{-15pt}
    \label{fig:sim-results}
\end{figure*}

\begin{figure}[th!]
\vspace{5pt}
    \centering
    \includegraphics[width=0.45\textwidth]{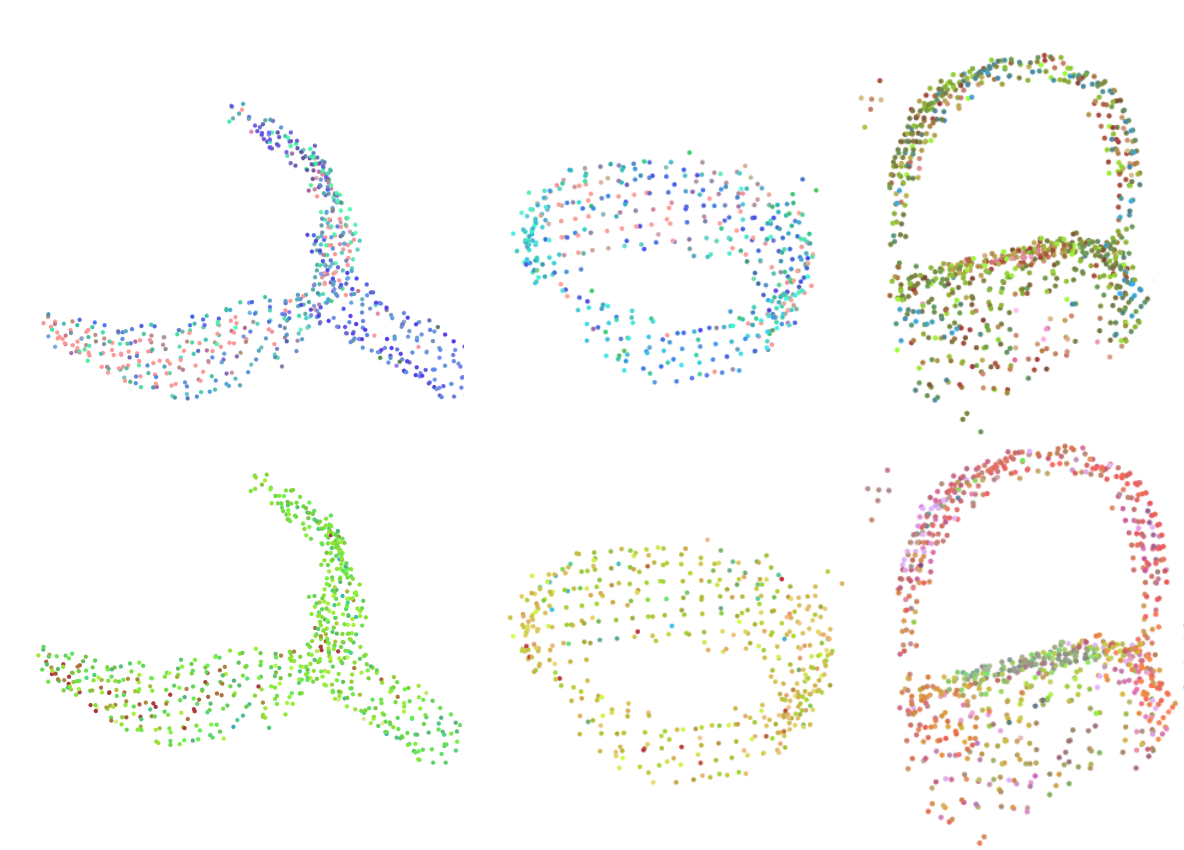}
    \caption{\textbf{Feature PCA Visualization for dustpan, bowl, headphone}. The first row represents feature distributions before applying our language alignment strategy, while the second row shows the results after alignment. The improved structure in the second row demonstrates how our method enhances feature consistency and smoothness, leading to better semantic coherence across different views. For a comparison with the teddy bear case, see Fig.\ref{fig:feature-distill}}
    \label{fig:vis_pca}
    \vspace{-15pt}
\end{figure}

\subsection{Simulation Experiments}
For evaluation, we record test scenes consisting of 12 YCB objects, with each scene containing three objects to create a mildly cluttered environment, one example shown in Fig.~\ref{fig:feature-distill-2}. Our method is evaluated against:
(1) our normal-based grasp sampler, (2) SparseDFF~\cite{wang2024sparsedff} and (3) F3RM~\cite{shen2023f3rm}. The initial normal-based initial grasp samples are evaluated directly for success rate without grasp optimization. For multi-view test cases, we capture 4 views for SparseDFF~\cite{wang2024sparsedff} and 50 views for F3RM~\cite{shen2023f3rm}. 
%For single-view test cases, we directly send a single view in SparseDFF~\cite{wang2024sparsedff} through the alignment network to assess its performance.
Since F3RM~\cite{shen2023f3rm} is not originally designed for dexterous hands, we only load the palm pose from their optimization, and we set a fixed target joint configuration that closes all fingers uniformly.

Every evaluated method generates 10 best grasps per object in each scene, totaling 120 grasps. These grasps are then sent through the real2sim pipeline for evaluation in Isaac Sim~\cite{NVIDIA2023IsaacSim}. In simulation, each grasp is executed by closing the fingers and enabling gravity after three contact points are established. The grasp is determined stable if the object remains securely held for at least 3s, same as~\cite{casas2024multigripper}.

% We use the fall time as the primary metric, categorizing it into 1s, 2s, and 3s, with longer times indicating higher quality grasps.%The number of successes at 0s serve as an indicator of successful object identification.

The simulation results are presented in Table.~\ref{tab:grasping_simulation_succ}.  
The normal-based grasp sampler has a significantly lower success rate compared to other methods, emphasizing the necessity of a few-shot learning framework. 
Our LensDFF outperforms SparseDFF~\cite{wang2024sparsedff} with $15.8\%$ and F3RM~\cite{shen2023f3rm} with $16.9\%$. A detailed grasp visualization is shown in Fig.~\ref{fig:sim-results}.  F3RM~\cite{shen2023f3rm} achieves slightly lower final results ($>3s$) than SparseDFF~\cite{wang2024sparsedff} but higher success rate at ($>0s$) demonstrate a good initial palm poses but bad finger configurations. 
Overall, the grasp success rates are somewhat low. One reason is that time-consuming collision avoidance is not applied to simulation experiments. Other reasons are possibly due to noisy real-world RealSense inputs and the strict grasp evaluation criteria in Isaac Sim.
%in multi-view cases possibly due to its use of 50 views and feature rendering, allowing it to generate a denser and high-quality feature field. However, dense view collections result in poor runtime efficiency.

% \begin{table}[t!]
% \vspace{5pt
% }\centering
% \ra{1.5}
% \caption{Average Success Rate and Run-time in Simulation {\color{red} change this to multigripper style}}
% % \vspace{15pt}
% \begin{center}
% \label{tab:grasping_simulation_succ} 
% \begin{adjustbox}{width=0.8\linewidth}
% \begin{tabular}{r|r r|r}
% Methods & Multi views & Single view &  \makecell{Run time (ms)}  \\ [0.0001cm] 
% \midrule
% Grasp Sampler & $2.5\%$ &  $0.4\%$ & - \\
% \midrule
% F3RM~\cite{shen2023f3rm} & $\textbf{24.17} \%$ & - & ~ 5 min \\
% % \midrule
% SparseDFF~\cite{wang2024sparsedff}  & $21.67\%$ & $30.00\%$ & 18s \\
% % SparseDFF~\cite{wang2024sparsedff} & $7.08\%$ & $20.83\%$ & $\textbf{30}$ \\
% LensDFF  & $22.50\%$ & $\textbf{40.83}\%$ & ~15s \\

% \end{tabular}
% \end{adjustbox}
% \vspace{-20pt}
% \end{center}
% \end{table}

\begin{table}[h]
    \centering
    \renewcommand{\arraystretch}{1.2}
    \caption{Average Success Rate and Run-time in Simulation}
    \vspace{3pt}
    \begin{adjustbox}{width=\linewidth}

    \begin{tabular}{l| cccc |c}
        \hline
        \textbf{Methods} & \multicolumn{4}{c|}{\textbf{Success Rate (\%)}} & \textbf{\#Grasps} \\
        & $>3s$ & $>2s$ & $>1s$ & $>0s$ & \\
        \hline
        Grasp Sampler      & 2.5 & 2.5 & 2.5 & 7.5  & 120 \\
        F3RM~\cite{shen2023f3rm}     & 23.9 & 24.1 & 24.6 & 82.6  & 120 \\
        SparseDFF~\cite{wang2024sparsedff}     & 25.0  & 25.0  & 25.0  & 53.3   & 120 \\
        LensDFF   & \textbf{40.8} & \textbf{40.8} & \textbf{41.7} & \textbf{85.0} & 120 \\

        \hline
    \end{tabular}
    \end{adjustbox}
    \vspace{-5pt}
\label{tab:grasping_simulation_succ} \end{table}

% \begin{table}[t!]
% \vspace{5pt
% }\centering
% \ra{1.5}
% \caption{Average Success Rate and Run-time in Simulation (new metrics)}
% \vspace{15pt}
% \begin{center}
% \label{tab:grasping_simulation_succ} 
% \begin{adjustbox}{width=0.8\linewidth}
% \begin{tabular}{r|rr|r}
% Methods & \makecell{Multi-view} & \makecell{Single-view} &  \makecell{Run time (ms)}  \\ [0.0001cm] 
% % \midrule
% % Heuristic & $20.9\%$ & $11.1\%$ & $3387$ \\
% \midrule
% F3rm~\cite{} & $- \%$ & $ - $ & $\textbf{5 min}$ \\
% SparseDFF~\cite{} & $17.51\%$ & $41.66\%$ & $\textbf{30}$ \\
% Ours~\cite{} & $4.0\%$ & $36.50\%$ & $1610$ \\

% \end{tabular}
% \end{adjustbox}
% \vspace{-1cm}
% \end{center}
% \end{table}

\subsection{Real-world Experiments} 
To validate the system's performance, we conduct real-world experiments. We first scan the scene and use the fused point cloud data to create an octomap for MoveIt collision avoidance.  Next, we perform grasp optimization and execute the best grasp on the physical robot. 
%This involves moving the arm to the target object, grasping it with the \hand, and lifting the object. We record a successful grasp if the object remains in the hand throughout the process.

Five YCB objects—mustard bottle, hammer, spam, blue screwdriver, and red mug—are selected from the 12-object test set and evaluated with 10 grasp attempts each in the real world, totaling 50 grasps. The results of these experiments are presented in Table.\ref{tab:grasping_real_succ}. Our LensDFF outperforms both baselines with $4\%$ and $10\%$ success rates.
Certain failure cases, especially for pinch and tripod grasps, where a high grasp pose accuracy is needed to ensure a stable grasp. One reason that the real-world success rate is higher than the simulation is that using MoveIt effectively filters unreachable and collided grasps.
We further compare the run time for each method. After the robot captures all views, the run-time is computed from object detection and feature extraction until computing the final grasps. Our run time is about 13s, including running SAM and Clip. The feature alignment takes only 70ms and grasp optimization for 10-11s. F3RM~\cite{shen2023f3rm}'s long run time partially results from its additional NeRF training. Both success rate and run time results showcase the effectiveness and efficiency of our approach.

\begin{table}[t!]
% \vspace{5pt}
\centering
\ra{1.5}
\caption{Average Success Rate in Real-World}
\vspace{-5pt}
\begin{center}
\label{tab:grasping_real_succ} 
\begin{adjustbox}{width=0.6\linewidth}
\begin{tabular}{r|r|r}
Methods & Success rate & Run time \\ [0.0001cm] 
% \midrule
% Heuristic & $20.9\%$ & $11.1\%$ & $3387$ \\
\midrule
F3RM~\cite{shen2023f3rm} & $ 60.0\% $ & 5 min \\
SparseDFF~\cite{wang2024sparsedff}  & $54.0\%$ & 16s \\
LensDFF & $\textbf{64.0}\%$ & 13s \\

\end{tabular}
\end{adjustbox}
\vspace{-1cm}
\end{center}
\end{table}

\subsection{Ablation Study}
To validate the design of LensDFF, we conduct ablation studies evaluating different feature alignment strategies and scene representations. In Table.~\ref{tab:ablation_align}, `No alignment' indicates a direct feature fusing. 
% Instead of LensDFF to align vision features with language, we also apply feature alignment with global vision features proposed in Conceptfusion~\cite{jata2023conceptfusion}. However, it seems even though our various views are targeting the same objects, the global vision features from the views are not good enough for alignment. 
`+ Language Feature Enhancement' simply projects vision features onto demo prompt language features without test-time feature alignment.  

\begin{table}[h]
\vspace{-5pt}
\centering
\ra{1.5}
\caption{Ablation study on different alignment strategies}
% \vspace{15pt}
\begin{center}
\label{tab:ablation_align} 
\begin{adjustbox}{width=0.7\linewidth}
\begin{tabular}{r|r|}
Methods & Success rate \\ [0.0001cm] 
% \midrule
\midrule
No alignment & $0\%$ \\
% Conceptfusion~\cite{jatavallabhula2023conceptfusio} & $5.83\%$\\
+ Language Feature Enhancement & $34.17\%$ \\
+ Test-Time Alignment (LensDFF) & $\textbf{40.83}\%$  \\
\end{tabular}
\end{adjustbox}
\vspace{-15pt}
\end{center}
\end{table}

\begin{table}[h]
\vspace{-5pt}
\centering
\ra{1.5}
\caption{Ablation study on different Demo/Test Representations}
% \vspace{15pt}
\begin{center}
\label{tab:ablation_representation} 
\begin{adjustbox}{width=0.7\linewidth}
\begin{tabular}{r|r|}
Methods & Success rate \\ [0.0001cm] 
% \midrule
\midrule
Single-View Demo + Single-View Test & 30.00\% \\
Multi-View Demo + Multi-View Test& 22.50\%\\
Multi-View Demo + Single-View Test & \textbf{40.83\%} \\
\end{tabular}
\end{adjustbox}
\vspace{-15pt}
\end{center}
\end{table}

Table.\ref{tab:ablation_representation} examines scene representations of multi-view and single-view. `Single-View Demo' treats each demo viewpoint separately, and each view is aligned in the same way as we treat test single view in LensDFF. The lower performance suggests that single-view demos lack sufficient information for optimal grasping. A lower performance suggests the limited info from single-view demos is not sufficient enough for grasp optimization. `Multi-View Test' fuses multiple test scene views. Interestingly, LensDFF performs better in single-view test than multi-view. This could be attributed to the cluttered scene where the target object may be occluded in certain views. In single-view test cases, unrecognized views will be skipped, and the next view will be chosen. In multi-view cases, unrecognized views trigger ``plane segmentation" to remove the table, which negatively affects the final DFF quality. These results highlight the importance of sparse multi-view demos and single-view test cases for achieving better grasping performance and efficiency. 
%===============================================================================
\section{Conclusion}

In this work, we propose LensDFF, Language-Enhanced Sparse Feature Distillation. This novel approach achieves efficient feature distillation from multiple 2D views onto 3D points using language feature alignment. Additionally, we incorporate grasp primitives into the demonstration collection process for a few-shot dexterous grasping framework, significantly improving grasping dexterity and grasp stability. Through our real2sim pipeline, we efficiently tune our framework and conduct extensive simulation and real-world experiments to validate its effectiveness. 
For future work, we aim to explore active learning for selecting more informative single-view observations.



% \section*{ACKNOWLEDGMENT}

% The preferred spelling of the word ÒacknowledgmentÓ in America is without an ÒeÓ after the ÒgÓ. Avoid the stilted expression, ÒOne of us (R. B. G.) thanks . . .Ó  Instead, try ÒR. B. G. thanksÓ. Put sponsor acknowledgments in the unnumbered footnote on the first page.

%%%%%%%%%%%%%%%%%%%%%%%%%%%%%%%%%%%%%%%%%%%%%%%%%%%%%%%%%%%%%%%%%%%%%%%%%%%%%%%%

% \bibliographystyle{IEEEtran}
% \bibliographystyle{plainnat}
% \bibliography{IEEEabrv, bibliography}

\bibliographystyle{IEEEtran}
{\scriptsize
\vspace{0.01 cm}
\bibliography{root}

\begin{thebibliography}{10}
\providecommand{\url}[1]{#1}
\csname url@samestyle\endcsname
\providecommand{\newblock}{\relax}
\providecommand{\bibinfo}[2]{#2}
\providecommand{\BIBentrySTDinterwordspacing}{\spaceskip=0pt\relax}
\providecommand{\BIBentryALTinterwordstretchfactor}{4}
\providecommand{\BIBentryALTinterwordspacing}{\spaceskip=\fontdimen2\font plus
\BIBentryALTinterwordstretchfactor\fontdimen3\font minus
  \fontdimen4\font\relax}
\providecommand{\BIBforeignlanguage}[2]{{%
\expandafter\ifx\csname l@#1\endcsname\relax
\typeout{** WARNING: IEEEtran.bst: No hyphenation pattern has been}%
\typeout{** loaded for the language `#1'. Using the pattern for}%
\typeout{** the default language instead.}%
\else
\language=\csname l@#1\endcsname
\fi
#2}}
\providecommand{\BIBdecl}{\relax}
\BIBdecl

\bibitem{ffhnet}
V.~Mayer, Q.~Feng, J.~Deng, Y.~Shi, Z.~Chen, and A.~Knoll, ``Ffhnet: Generating
  multi-fingered robotic grasps for unknown objects in real-time,'' in
  \emph{2022 International Conference on Robotics and Automation (ICRA)}, 2022,
  pp. 762--769.

\bibitem{zhang2024dexgraspnet20}
J.~Zhang, H.~Liu, D.~Li, X.~Yu, H.~Geng, Y.~Ding, J.~Chen, and H.~Wang,
  ``Dexgraspnet 2.0: Learning generative dexterous grasping in large-scale
  synthetic cluttered scenes,'' 2024.

\bibitem{feng2024ffhflowflow}
Q.~Feng, J.~Feng, Z.~Chen, R.~Triebel, and A.~Knoll, ``Ffhflow: A flow-based
  variational approach for learning diverse dexterous grasps with shape-aware
  introspection,'' 2024.

\bibitem{weng2024dexdiffuser}
Z.~Weng, H.~Lu, D.~Kragic, and J.~Lundell, ``Dexdiffuser: Generating dexterous
  grasps with diffusion models,'' 2024.

\bibitem{liu2020deepdifferentiablegrasp}
M.~Liu, Z.~Pan, K.~Xu, K.~Ganguly, and D.~Manocha, ``Deep differentiable grasp
  planner for high-dof grippers,'' 2020.

\bibitem{wang2023dexgraspnet}
R.~Wang, J.~Zhang, J.~Chen, Y.~Xu, P.~Li, T.~Liu, and H.~Wang, ``Dexgraspnet: A
  large-scale robotic dexterous grasp dataset for general objects based on
  simulation,'' 2023.

\bibitem{radford2021clip}
A.~Radford, J.~W. Kim, C.~Hallacy, A.~Ramesh, G.~Goh, S.~Agarwal, G.~Sastry,
  A.~Askell, P.~Mishkin, J.~Clark, G.~Krueger, and I.~Sutskever, ``Learning
  transferable visual models from natural language supervision,'' 2021.

\bibitem{kirillov2023sam}
A.~Kirillov, E.~Mintun, N.~Ravi, H.~Mao, C.~Rolland, L.~Gustafson, T.~Xiao,
  S.~Whitehead, A.~C. Berg, W.-Y. Lo, P.~Dollár, and R.~Girshick, ``Segment
  anything,'' 2023.

\bibitem{oquab2024dinov2}
M.~Oquab, T.~Darcet, T.~Moutakanni, H.~Vo, M.~Szafraniec, V.~Khalidov,
  P.~Fernandez, D.~Haziza, F.~Massa, A.~El-Nouby, M.~Assran, N.~Ballas,
  W.~Galuba, R.~Howes, P.-Y. Huang, S.-W. Li, I.~Misra, M.~Rabbat, V.~Sharma,
  G.~Synnaeve, H.~Xu, H.~Jegou, J.~Mairal, P.~Labatut, A.~Joulin, and
  P.~Bojanowski, ``Dinov2: Learning robust visual features without
  supervision,'' 2024.

\bibitem{shen2023f3rm}
W.~Shen, G.~Yang, A.~Yu, J.~Wong, L.~P. Kaelbling, and P.~Isola, ``Distilled
  feature fields enable few-shot language-guided manipulation,'' 2023.

\bibitem{rashid2023lerftogo}
A.~Rashid, S.~Sharma, C.~M. Kim, J.~Kerr, L.~Chen, A.~Kanazawa, and
  K.~Goldberg, ``Language embedded radiance fields for zero-shot task-oriented
  grasping,'' 2023.

\bibitem{wang2024sparsedff}
Q.~Wang, H.~Zhang, C.~Deng, Y.~You, H.~Dong, Y.~Zhu, and L.~Guibas,
  ``Sparsedff: Sparse-view feature distillation for one-shot dexterous
  manipulation,'' 2024.

\bibitem{zheng2024gaussiangrasper}
Y.~Zheng, X.~Chen, Y.~Zheng, S.~Gu, R.~Yang, B.~Jin, P.~Li, C.~Zhong, Z.~Wang,
  L.~Liu, C.~Yang, D.~Wang, Z.~Chen, X.~Long, and M.~Wang, ``Gaussiangrasper:
  3d language gaussian splatting for open-vocabulary robotic grasping,'' 2024.

\bibitem{ji2024graspsplats}
M.~Ji, R.-Z. Qiu, X.~Zou, and X.~Wang, ``Graspsplats: Efficient manipulation
  with 3d feature splatting,'' 2024.

\bibitem{kobayashi2022decomposingnerf}
S.~Kobayashi, E.~Matsumoto, and V.~Sitzmann, ``Decomposing nerf for editing via
  feature field distillation,'' 2022.

\bibitem{kerr2023lerf}
J.~Kerr, C.~M. Kim, K.~Goldberg, A.~Kanazawa, and M.~Tancik, ``Lerf: Language
  embedded radiance fields,'' 2023.

\bibitem{wang2024neuralattentionfield}
Q.~Wang, C.~Deng, T.~G.~W. Lum, Y.~Chen, Y.~Yang, J.~Bohg, Y.~Zhu, and
  L.~Guibas, ``Neural attention field: Emerging point relevance in 3d scenes
  for one-shot dexterous grasping,'' 2024.

\bibitem{ciocarlie2009hand}
M.~T. Ciocarlie and P.~K. Allen, ``Hand posture subspaces for dexterous robotic
  grasping,'' \emph{The International Journal of Robotics Research}, vol.~28,
  no.~7, pp. 851--867, 2009.

\bibitem{Arbib2008FromGT}
M.~A. Arbib, ``From grasp to language: Embodied concepts and the challenge of
  abstraction,'' \emph{Journal of Physiology-Paris}, vol. 102, pp. 4--20, 2008.

\bibitem{IVERSONJANAM2010Dlia}
J.~M. IVERSON, ``\BIBforeignlanguage{eng}{Developing language in a developing
  body: the relationship between motor development and language development},''
  \emph{\BIBforeignlanguage{eng}{Journal of child language}}, vol.~37, no.~2,
  pp. 229--261, 2010.

\bibitem{mildenhall2020nerf}
B.~Mildenhall, P.~P. Srinivasan, M.~Tancik, J.~T. Barron, R.~Ramamoorthi, and
  R.~Ng, ``Nerf: Representing scenes as neural radiance fields for view
  synthesis,'' 2020.

\bibitem{kerbl20233dgaussiansplatt}
B.~Kerbl, G.~Kopanas, T.~Leimkühler, and G.~Drettakis, ``3d gaussian splatting
  for real-time radiance field rendering,'' 2023.

\bibitem{simeonov2021neuraldescriptorfield}
A.~Simeonov, Y.~Du, A.~Tagliasacchi, J.~B. Tenenbaum, A.~Rodriguez, P.~Agrawal,
  and V.~Sitzmann, ``Neural descriptor fields: Se(3)-equivariant object
  representations for manipulation,'' 2021.

\bibitem{dai2023graspnerf}
Q.~Dai, Y.~Zhu, Y.~Geng, C.~Ruan, J.~Zhang, and H.~Wang, ``Graspnerf:
  Multiview-based 6-dof grasp detection for transparent and specular objects
  using generalizable nerf,'' 2023.

\bibitem{Lei2017cshape}
Q.~Lei, J.~Meijer, and M.~Wisse, ``Fast c-shape grasping for unknown objects,''
  in \emph{2017 IEEE International Conference on Advanced Intelligent
  Mechatronics (AIM)}, 2017, pp. 509--516.

\bibitem{lu2017grasp}
Q.~Lu, K.~Chenna, B.~Sundaralingam, and T.~Hermans, ``Planning multi-fingered
  grasps as probabilistic inference in a learned deep network,'' in
  \emph{Int’l Symp. on Robotics Research}, 2017.

\bibitem{Lu2019Reconstruct}
M.~V. der Merwe, Q.~Lu, B.~Sundaralingam, M.~Matak, and T.~Hermans, ``Learning
  continuous 3d reconstructions for geometrically aware grasping,''
  \emph{CoRR}, vol. abs/1910.00983, 2019.

\bibitem{feng2024dexgangrasp}
Q.~Feng, D.~S.~M. Lema, M.~Malmir, H.~Li, J.~Feng, Z.~Chen, and A.~Knoll,
  ``Dexgangrasp: Dexterous generative adversarial grasping synthesis for
  task-oriented manipulation,'' 2024.

\bibitem{wei2024funcgrasp}
W.~Wei, P.~Wang, S.~Wang, Y.~Luo, W.~Li, D.~Li, Y.~Huang, and H.~Duan,
  ``Learning human-like functional grasping for multifinger hands from few
  demonstrations,'' \emph{IEEE Transactions on Robotics}, vol.~40, pp.
  3897--3916, 2024.

\bibitem{Li2022hgcnet}
Y.~Li, W.~Wei, D.~Li, P.~Wang, W.~Li, and J.~Zhong, ``Hgc-net: Deep
  anthropomorphic hand grasping in clutter,'' in \emph{2022 International
  Conference on Robotics and Automation (ICRA)}, 2022, pp. 714--720.

\bibitem{chen2022isagrasp}
Z.~Q. Chen, K.~V. Wyk, Y.-W. Chao, W.~Yang, A.~Mousavian, A.~Gupta, and D.~Fox,
  ``Learning robust real-world dexterous grasping policies via implicit shape
  augmentation,'' 2022.

\bibitem{jeremy2016oneshot}
M.~Kopicki, R.~Detry, M.~Adjigble, R.~Stolkin, A.~Leonardis, and J.~L. Wyatt,
  ``One-shot learning and generation of dexterous grasps for novel objects,''
  \emph{The International Journal of Robotics Research}, vol.~35, no.~8, pp.
  959--976, 2016.

\bibitem{ravi2024sam2}
N.~Ravi, V.~Gabeur, Y.-T. Hu, R.~Hu, C.~Ryali, T.~Ma, H.~Khedr, R.~Rädle,
  C.~Rolland, L.~Gustafson, E.~Mintun, J.~Pan, K.~V. Alwala, N.~Carion, C.-Y.
  Wu, R.~Girshick, P.~Dollár, and C.~Feichtenhofer, ``Sam 2: Segment anything
  in images and videos,'' 2024.

\bibitem{zhou2019continuity}
Y.~Zhou, C.~Barnes, J.~Lu, J.~Yang, and H.~Li, ``On the continuity of rotation
  representations in neural networks,'' in \emph{Proceedings of the IEEE/CVF
  Conference on Computer Vision and Pattern Recognition}, 2019, pp. 5745--5753.

\bibitem{zhong2021regionclip}
Y.~Zhong, J.~Yang, P.~Zhang, C.~Li, N.~Codella, L.~H. Li, L.~Zhou, X.~Dai,
  L.~Yuan, Y.~Li, and J.~Gao, ``Regionclip: Region-based language-image
  pretraining,'' 2021.

\bibitem{calli2015ycb}
B.~Calli, A.~Singh, A.~Walsman, S.~Srinivasa, P.~Abbeel, and A.~M. Dollar,
  ``The ycb object and model set: Towards common benchmarks for manipulation
  research,'' in \emph{2015 international conference on advanced robotics
  (ICAR)}.\hskip 1em plus 0.5em minus 0.4em\relax IEEE, 2015, pp. 510--517.

\bibitem{liu2008hithand}
H.~Liu, K.~Wu, P.~Meusel, N.~Seitz, G.~Hirzinger, M.~Jin, Y.~Liu, S.~Fan,
  T.~Lan, and Z.~Chen, ``Multisensory five-finger dexterous hand: The dlr/hit
  hand ii,'' in \emph{2008 IEEE/RSJ International Conference on Intelligent
  Robots and Systems}, 2008, pp. 3692--3697.

\bibitem{wen2024foundationpose}
B.~Wen, W.~Yang, J.~Kautz, and S.~Birchfield, ``Foundationpose: Unified 6d pose
  estimation and tracking of novel objects,'' 2024.

\bibitem{NVIDIA2023IsaacSim}
\BIBentryALTinterwordspacing
NVIDIA, ``Nvidia isaac sim: Robotics simulation and synthetic data,'' Online,
  2023. [Online]. Available: \url{developer.nvidia.com/isaac-sim}
\BIBentrySTDinterwordspacing

\bibitem{casas2024multigripper}
L.~F. Casas, N.~Khargonkar, B.~Prabhakaran, and Y.~Xiang, ``Multigrippergrasp:
  A dataset for robotic grasping from parallel jaw grippers to dexterous
  hands,'' 2024.

\end{thebibliography}
}

\end{document}